\documentclass[runningheads]{llncs}
\usepackage{graphicx}

\usepackage{tikz}
\usepackage{comment}
\usepackage{amsmath,amssymb} 
\usepackage{color}
\usepackage{caption} 

\usepackage[accsupp]{axessibility}  

\usepackage{booktabs}
\usepackage{pifont}
\usepackage{subcaption}
\usepackage{gensymb}
\usepackage{makecell}

\usepackage[capitalize]{cleveref}

\crefname{section}{Sec.}{Secs.}
\Crefname{section}{Section}{Sections}
\Crefname{table}{Table}{Tables}
\crefname{table}{Tab.}{Tabs.}

\newcommand{\figref}[1]{\figurename~\ref{#1}}

\newcommand{\etal}{\textit{et al}. }

\newcommand\blfootnote[1]{%
  \begingroup
  \renewcommand\thefootnote{}\footnote{#1}%
  \addtocounter{footnote}{-1}%
  \endgroup
}

\begin{document}
\pagestyle{headings}
\mainmatter
\def\ECCVSubNumber{3500}  

\title{Weakly-Supervised Stitching Network for Real-World Panoramic Image Generation} 

\titlerunning{Weakly-Supervised Stitching for Panoramic Image Generation}
%
\author{Dae-Young Song\inst{1} \and
Geonsoo Lee\inst{1} \and
HeeKyung Lee\inst{2} \and
Gi-Mun Um\inst{2} \and
Donghyeon Cho*\inst{1}}
\authorrunning{D.-Y. Song et al.}
%
\institute{Chungnam National University, Daejeon, South Korea \\
\email{\{201501747.o,geonsoo.o,cdh12242\}@cnu.ac.kr} \and
Electronics and Telecommunication Research Institute, Daejeon, South Korea\\
\email{\{lhk95,gmum\}@etri.re.kr}}
\maketitle

\begin{abstract}
Recently, there has been growing attention on an end-to-end deep learning-based stitching model.
However, the most challenging point in deep learning-based stitching is to obtain pairs of input images with a narrow field of view and ground truth images with a wide field of view captured from real-world scenes.
To overcome this difficulty, we develop a weakly-supervised learning mechanism to train the stitching model without requiring genuine ground truth images.
In addition, we propose a stitching model that takes multiple real-world fisheye images as inputs and creates a 360$^{\circ}$ output image in an equirectangular projection format.
In particular, our model consists of color consistency corrections, warping, and blending, and is trained by perceptual and SSIM losses.
The effectiveness of the proposed algorithm is verified on two real-world stitching datasets.
\end{abstract}

\blfootnote{ * Corresponding author.  \\ Project page is at \textcolor{magenta}{\tt \url{https://eadcat.github.io/WSSN}}}

\section{Introduction}
Image stitching is a task that combines multiple images obtained from different viewpoints to generate a single panoramic image with a larger field of view (FOV). 
By exploiting this advantage, image stitching technique can be used in various applications such as street view service, virtual reality~\cite{li2019cross}, video surveillance~\cite{he2016parallax}, and Mars exploration~\cite{coates2017pancam}.
Traditional stitching proceeds in the order of feature point extraction, feature matching, homography estimation, warping, and blending.
For instance, Brown and Lowe~\cite{brown2007automatic} proposed an automatic stitching method that finds correspondence of feature points using SIFT~\cite{lowe2004distinctive}, estimates global homography by RANSAC~\cite{fischler1981ransac}, aligns two images using estimated homography, and combines them by multi-band blending.
Since then, a lot of following methods have been developed for creating high-quality panoramic images, and the main research issue of them is to deal with parallax distortion caused by depth differences.
To overcome parallax distortion, spatially varying multi-homography estimation methods~\cite{gao2011constructing,lin2011smoothly,gao2013seam,xu2017wide,herrmann2018robust,lee2020warping} and non-uniform mesh-based warping methods~\cite{zaragoza2013projective,he2013rectangling,zhang2014parallax} have been introduced.
Additionally, another challenge to consider in real-world stitching is to overcome visually unpleasant artifacts such as structural distortion and the difference in overall tones between input images.
To seamlessly combine input images without suffering these distortions, constraints on line and structure can be explicitly included in the stitching algorithm~\cite{xiang2018image,liao2019single,jia2021leveraging}, and color consistency correction can be applied to input images based on a parametric color model~\cite{Doutre_2009_ICIP,Xia_2017_ICCV}.
However, the aforementioned approaches depend on the performance of the algorithm that estimates the correspondence of feature points between the input images.
%
Therefore, when the overlapping area between input images is too small or there are many repetitive patterns, feature matching becomes challenging, resulting in parallax distortion and visually unpleasant artifacts, or stitching itself may fail. In other words, the success rate of stitching depends on the performance of the matching algorithm.

\begin{figure}[t]
   \centering
   \includegraphics[width=1.0\linewidth]{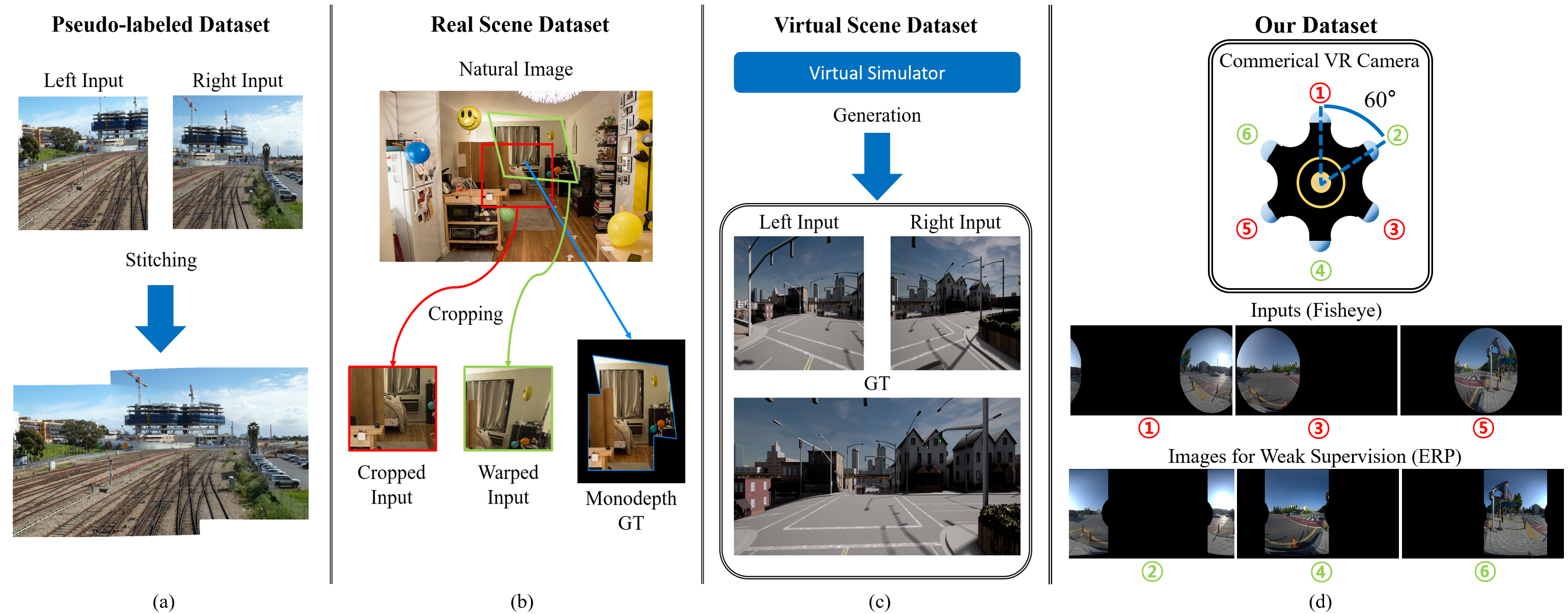}
   \caption{Comparison of existing stitching dataset for training. (a) The pseudo-labeled dataset is constructed by existing stitching methods. (b) The real scene dataset is generated by cropping with global homography. (c) Virtual scene dataset with a simulator. (d) Our dataset for weak supervisions.
   }
   \label{fig:teaser}
\end{figure}

Recently, the limitations of these traditional approaches have been solved by the CNN-based feature matching technique~\cite{sarlin2020superglue} and deep homography estimation methods~\cite{detone2016deep,nguyen2018unsupervised,wang2019self,le2020deep,zhang2020content}.
Furthermore, researches on modeling the entire stitching process as a single pipeline based on neural network are being introduced~\cite{shen2019real,li2019attentive,lai2019video,nie2020view,nie2021unsupervised,song2021end,dai2021edge}.
Unlike feature matching and homography estimation, it is difficult to construct the inputs-GT pairs for training the end-to-end deep stitching model.
To get inputs-GT pairs, Shen~\etal~\cite{shen2019real} built a unique hardware system that can capture the real-world scene with fixed viewpoints, but it cannot contain dynamic objects due to its systemic limitation.
In addition, there were several efforts~\cite{li2019cross,li2019attentive,dai2021edge} to make pseudo GT labels by applying existing stitching methods to real-world images.
However, pseudo GT labels may be sensitive to the methods used to create them.
In~\cite{nie2020view}, inputs-GT pairs were constructed by cropping sub-images from natural images with random geometric transformations.
However, it cannot cover various depths because it is a crop-based method.
To handle multiple depth layers and moving objects, there have been studies~\cite{lai2019video,song2021end} using a virtual simulator such as CARLA~\cite{dosovitskiy2017carla} to generate inputs-GT pairs.
However, it is difficult to use stitching models trained with synthetic datasets on real-world images without the help of domain adaptation.
In summary, constructing real inputs-GT pairs that take the depth of the scene into account for training an end-to-end stitching model is a very challenging problem.
Therefore, in this paper, we present a weakly-supervised learning method for training a deep stitching model.
To this end, we use a commercial camera to capture six fisheye images uniformly rotated at 60$^{\circ}$ intervals.
We use half of the captured images ($0^{\circ}$, $120^{\circ}$, $240^{\circ}$) as inputs and the other remaining images ($60^{\circ}$, $180^{\circ}$, $300^{\circ}$) as weak supervisions.
Note that all images are captured simultaneously, thus dynamic scenes and objects can be covered in our dataset.
Then, we introduce a novel mechanism to train an end-to-end stitching model using our dataset.
Meanwhile, we develop a deep stitching model that performs color consistency corrections, warping, and blending.
In addition, our model and training mechanism can be applied to the existing pseudo GT-based dataset~\cite{li2019cross}.
Comparisons of training datasets for the stitching model are shown in~\figref{fig:teaser}.
Our contributions can be summarized as follows:
\begin{itemize}
  \item We introduce a novel weak-supervised method for training a stitching network to create real-world $360^{\circ}$ panoramic images.
  \item Our stitching model can effectively deal with parallax distortion due to depth differences as well as inconsistent colors between input images.
  \item We provide a variety of ablation studies, including the results of training the proposed model using the existing CROSS dataset~\cite{li2019cross}.
\end{itemize}

\section{Related Works}
\label{sec:related_work}
In this section, we review both traditional stitching methods and recent deep learning-based stitching methods.
\subsection{Traditional Stitching Methods}
After Brown and Lowe~\cite{brown2007automatic} introduced an automatic stitching method using SIFT feature~\cite{lowe2004distinctive}, RANSAC~\cite{fischler1981ransac}, and multi-band blending, lots of follow-up studies addressing various issues have been introduced.

\noindent\textbf{Parallax distortion.} To handle multiple depth layers in the scene, Gao~\etal~\cite{gao2011constructing} proposed a method that estimates dual homography for two separate regions: ground plane and distant plane.
Lin~\etal~\cite{lin2011smoothly} introduced a spatially varying affine field to adaptively align pixels.
Zaragoza~\etal~\cite{zaragoza2013projective} proposed as-projective-as-possible (APAP) image stitching based on moving direct linear transformation (Moving DLT) for allowing local non-projective deviations.
In~\cite{zhang2014parallax}, input images are aligned by estimated homography, then content-preserving warping is applied to solve local parallax distortion.
However, the existing methods have problems such as perspective distortion when stitching multiple images.
Perazzi~\etal~\cite{perazzi2015panoramic} proposed a video stitching technique using multiple scenes from unstructured camera arrays. It deals well with parallax and perspective distortions, but takes a long time due to its large computational complexity.
In addition, seam-driven image stitching that finds the best homography based on the quality of seam-cut was introduced in~\cite{gao2013seam}, while seam-guided local alignment methods were proposed in~\cite{lin2016seagull}.
Herrmann~\etal~\cite{herrmann2018robust} proposed a robust stitching method that generates multiple registrations and combines them using Markov random field (MRF) with energy terms discouraging duplication and tearing effects.
Recently, Lee and Sim~\cite{lee2020warping} introduced a novel concept of warping residual to deal with large parallax using locally optimal warping.
For a similar goal, our network includes warping operations that take global and local information into account.

\noindent\textbf{Visually unpleasant distortion.}
In human visual perception, distortion tends to be particularly noticeable on thin objects such as lines or curves.
Xiang~\etal~\cite{xiang2018image} proposed a line-guided local warping method with a global similarity constraint to overcome projective distortions.
Liao~\etal~\cite{liao2019single} presented two single-perspective warpings consisting of parametric warping and mesh-based warping for enhancing the naturalness of stitched images.
Jia~\etal~\cite{jia2021leveraging} presented a structure-preserving method based on line-guided warping and line-point constraint.
Also, methods exploiting semantic information about pedestrians~\cite{flores2010removing},  faces~\cite{ozawa2012human}, human perception~\cite{li2018perception} and objects~\cite{herrmann2018object} were introduced for natural stitching.
%
%
%
%
%
%
In addition, color and tone differences between input images are noticeable distortions. Especially in the near of the seam lines, the distortion becomes more prominent.
Doutre and Nasiopoulos~\cite{Doutre_2009_ICIP} proposed a method that corrects differences in color between images using simple linear regression.
A more advanced color consistency correction method using convex quadratic programming for the stitching problem is proposed in~\cite{Xia_2017_ICCV}. 
To satisfy human visual perception, we utilize perceptual loss~\cite{ledig2017photo} in the training step and include color correction operation in our stitching model.

\subsection{Deep Learning-based Stitching Methods}
To train a deep stitching model, it is necessary to construct pairs of input images with a narrow FOV and a GT image with a wide FOV.
Shen~\etal~\cite{shen2019real} built the hardware system with a flat mirror to create the dataset and trained a stitching model using the constructed dataset.
Since it is not practical to use a specialized camera, there have been several studies that make inputs-GT pair as follows.

\noindent\textbf{Dataset with pseudo GT.}
Li~\etal~\cite{li2019cross} captured 4 fisheye images taken by lenses rotated at 90$^{\circ}$ intervals. 
Then, two images facing opposite directions are used as inputs, and the stitched image using the other two images is used as a pseudo GT image.
To create the stitched images, a method with the highest mean opinion score (MOS) among existing stitching methods is used for each image.
Using this dataset, Li~\etal~\cite{li2019attentive} introduced an attentive deep stitching approach consisting of two modules for deformation and resolution.
Similarly, Dai~\etal~\cite{dai2021edge} generated pseudo GT images using existing stitching methods and used them to train an edge-guided composition network.
An example of pseudo GT is illustrated in~\figref{fig:teaser}-(a).
However, pseudo GT labels are sensitive to the methods used to generate them. 

\noindent\textbf{Dataset with only global homography.}
Nie~\etal~\cite{nie2020view} presented a deep learning-based view-free stitching model consisting of global homography estimation, structure stitching, and content revision.
For the training, they constructed inputs-GT pairs using natural images such as the COCO dataset~\cite{lin2014microsoft} as shown in~\figref{fig:teaser}-(b).
Specifically, given an image, two sub-images having overlapping regions are extracted, then geometric transformation is applied to one of them. 
Thus, these two sub-images have different perspectives, and their geometric relationship can be modeled by a global homography.
These two sub-images are used as inputs to the stitching model while an image containing both sub-images is used as GT.
However, the problem with their dataset is that depth is not considered when generating two input images.
It means that a single depth layer is assumed, which is unrealistic in the real-world scenario. 
As a result, there is a limitation to stitching images containing scenes with multiple depth layers.
Furthermore, parallax distortion caused by depth differences that may occur in the real-world environment cannot be dealt with.
Recently, Nie~\etal~\cite{nie2021unsupervised} proposed an unsupervised learning method for a view-free stitching model composed of coarse alignment and image reconstruction.
However, the unsupervised coarse alignment module is performed by a global homography.
Thus, parallax distortion induced by depth difference still causes visual artifacts, even though the image reconstruction module enhances the quality of the output image.

\noindent\textbf{Dataset using virtual simulator.}
There have been several studies that train a stitching model by using inputs-GT pairs generated from a virtual simulator such as CARLA~\cite{dosovitskiy2017carla}. 
Since it is possible to control camera configuration and the scene in the virtual space, depth information can be included in the relationship between inputs as shown in~\figref{fig:teaser}-(c).
Thus, with these virtual datasets, parallax distortion due to depth differences can be covered.
Using the virtual dataset, Lai~\etal~\cite{lai2019video} proposed a pushbroom stitching network that estimates flow maps in fixed view, and Song~\etal~\cite{song2021end} developed an end-to-end virtual image stitching network via multi-homography estimation.
However, these stitching models trained with virtual dataset has limitations in applying them to real-world images, and additional techniques such as domain adaptation may be required.
In summary, it is hard to obtain real-world datasets that take the depth information of the scenes into account.
In this paper, we use real-world images captured at different viewpoints themselves as inputs to the stitching model.
In this case, there are no GT images with a wide FOV, thus we propose a new mechanism for training the stitching model.

\section{Approach}
\label{sec:stitching_model}
In this section, we first describe the procedure of generating training data using real-world fisheye images.
Then, we define the problem setup for creating a 360$^{\circ}$ panoramic image and introduce an architecture of the proposed stitching model to solve the defined problems. 
Finally, loss functions for training the proposed stitching model are explained.

\subsection{Dataset Preparation}
\label{sec:dataset}
To construct the training data for learning the real-world stitching model, we use a commercial VR camera called Kandao Obsidian R~\cite{kandao} to acquire fisheye images.
It can capture six fisheye images simultaneously using six lenses rotated at 60$^{\circ}$ intervals.
We use three fisheye images rotated by $0^{\circ}$, $120^{\circ}$, $240^{\circ}$ as inputs to our stitching model while the remaining three images rotated by $60^{\circ}$, $180^{\circ}$, $300^{\circ}$ are utilized as weak supervisions. 
As shown in~\figref{fig:teaser}-(d), overlapping areas between the input images correspond to the central regions of the images for weak supervisions.
Therefore, when training a stitching model using three input images, the remaining three images can be used as weak supervisions.
Input images are used as themselves whereas pre-processing is applied on images for weak supervisions.
Two types of pre-processing are performed on images for weak supervisions as follows.

\noindent{\textbf{Geometric calibration.}}
We represent the GT $360^{\circ}$ panoramic image in equirectangular projection (ERP) format.
Since there are no genuine GT images in our setup, it is required to register images for weak supervisions as much as possible in the GT format in advance.
Therefore, we transform images for weak supervisions into ERP coordinates.
To this end, we perform geometric calibration for fisheye cameras to compute intrinsic and extrinsic parameters by utilizing NVIDIA VRWorks 360 Video SDK~\cite{vrworks}.
As shown in~\figref{fig:teaser}-(d), three fisheye images for weak supervisions are well projected on ERP coordinates.

\noindent{\textbf{Color consistency correction.}}
Multi-view images captured in a real-world environment may have different color tones.
To utilize three images for weak supervisions as GT, the color tones of them should be matched consistently.
Therefore, we correct the color consistency of three images for weak supervisions in advance.
We use the polynomial curve mapping function to convert the color values of two images (called query image) to the those of remaining one image (called reference image) as
\begin{equation}
    \Bar{x}=ax^2+bx+c,
    \label{2nd-polynomial}
\end{equation}
where $x$ is the original pixel value in query images, $\Bar{x}$ is the corrected pixel value, and $a$, $b$, and $c$ are learnable parameters of the polynomial model, respectively.
We estimate $a$, $b$, and $c$ as follows.
First, we find correspondences between query and reference images using SuperGlue~\cite{sarlin2020superglue}, then extract patches centered at matched points from both images.
Then, we minimize mean squared error (MSE) between corrected patches from query images by~\eqref{2nd-polynomial} and extracted patches from the reference images.
Then, we obtain three images with consistent color tones in ERP format by using~\eqref{2nd-polynomial} with the learned $a$, $b$, and $c$.
These three images are used for weak supervisions to train our stitching model.
%
 \begin{figure*}[t]
   \centering
   \includegraphics[width=0.8\linewidth]{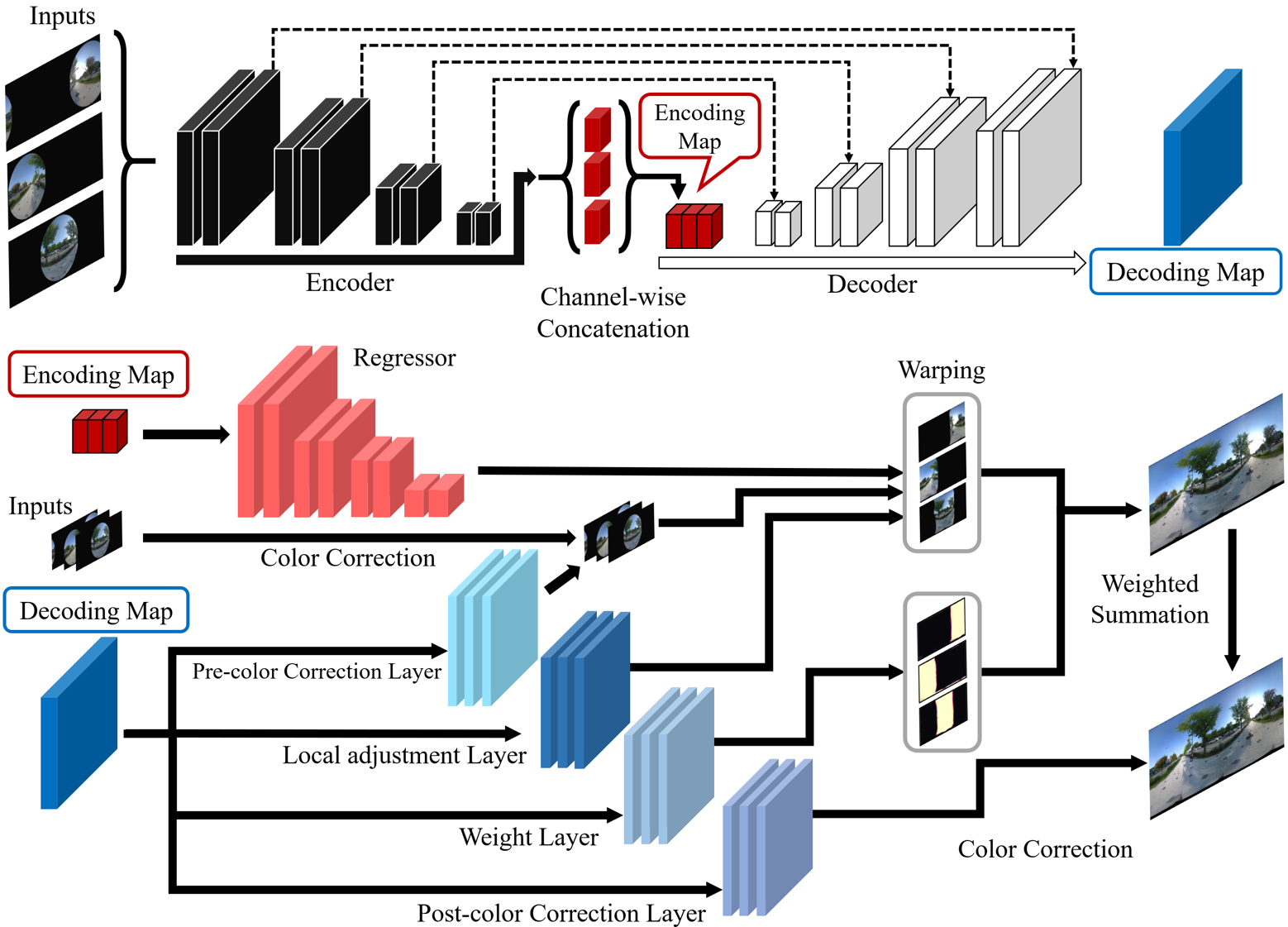}
   \caption{The entire pipeline of our stitching model. Our model takes $N$ inputs and produces global warping maps, pre- and post-color correction maps, local adjustment maps, and weight maps. After extracting an encoding map~(red) and a decoding map~(blue), a final warping map $U_{n}$ is created by adding a global warping map and a local adjustment map, and then the input is warped with $U_{n}$. All warped images are weighted by weight maps and merged into a panorama. 
   Color correction is applied once before warping and once after weighted summation using color correction maps.}
   \label{fig:overview}
 \end{figure*}
\subsection{Problem Definition}
In this paper, we aim to create a 360$^{\circ}$ panoramic image by stitching $N$ adjacent images taken with fisheye lenses rotated at different angles.
Our stitching model $\mathbf{S}(\cdot)$ takes $N$ fisheye images $I_{n}$ as inputs and generates a pre-color correction map $C_{n}^{pre}$, a global warping map $G_{n}$, a local warping adjustment map $L_{n}$, and a weight map $W_{n}$ for each input image. It also produces a post-color correction map $C^{post}$ for an input pair of $N$ images.
It is defined as
\begin{equation}
   \mathbf{S}(I_{1},...,I_{N} ;\theta) \rightarrow  (C_{n}^{pre}, G_{n}, L_{n}, W_{n}, C^{post}),
  \label{eq:problem}
 \end{equation}
where $\theta$ and $n$ are learnable parameters of $\mathbf{S}(\cdot)$ and the index of input images, respectively. 
In our experiment, $N$ is 3, the vertical FOV of each fisheye image is $185^{\circ}$, and the lens for each input is rotated $60^{\circ}$ from each other.
The panoramic image is created by applying all estimates from the stitching model in~\eqref{eq:problem} to the input fisheye images.
%
\subsection{Architecture}
Our stitching model $\mathbf{S}(\cdot)$ is composed of an encoder $\mathbf{E}(\cdot)$, a regressor $\mathbf{R}(\cdot)$, and a decoder $\mathbf{D}(\cdot)$ as illustrated in~\figref{fig:overview}. 
The role and details of each component are described as follows.

\noindent\textbf{Encoder.}
In our stitching model, there are $N$ encoders to extract visual features $f_{n}$ of each input fisheye image as
\begin{equation}
   f_{n} =  \mathbf{E}(I_{n};\theta_{e}),
  \label{eq:encoder}
 \end{equation}
where $I_{n}$ is one of the input fisheye images and $\theta_{e}$ is the learnable parameters of the encoder.
Our encoder consists of a series of convolutional layers, batch normalization layers, and ELU activations~\cite{clevert2015fast}.
Learnable parameters of each encoder are shared.
Visual features extracted from each input image are concatenated along the channel axis and used as input for a regressor and a decoder.

\noindent\textbf{Regressor.}
The purpose of the regressor is to find affine transformation matrices that can warp the pixel values of each input image to the pixel coordinates of the output image globally.
The regressor takes the visual features $f_{n}$ as input and generates affine matrices $A_{n}$ as follows.
\begin{equation}
   A_{n} =  \mathbf{R}(f_{n};\theta_{r}),
  \label{eq:regressor}
 \end{equation}
where $\theta_{r}$ is the learnable parameters of the regressor. 
Using the estimated affine matrices, a global warping map $G_{n}$ for each input image is created. 
The global warping map contains $x$- and $y$-direction information on where the pixels of the input image are moved to the coordinates of the output pixels.
Global warping can be viewed as global registration by a single homography.

\noindent\textbf{Decoder.}
Except for the global warping map $G_{n}$, the remaining components in~\eqref{eq:problem} needed to make the final output are generated by the decoder as
\begin{equation}
   \mathbf{D}(f_{n} ;\theta_{d}) \rightarrow  (C_{n}^{pre}, L_{n}, W_{n}, C^{post}),
  \label{eq:decoder}
 \end{equation}
where $\theta_{d}$ is the learnable parameters of the decoder.
Specifically, the decoder consists of a shared decoder and four private decoders for each output component.
The shared decoder takes the visual features obtained from the encoder as inputs and generates shared features.
Shared features are passed as input to each private decoder to create each output component.
A shared decoder consists of a series of convolutional blocks and upsampling layers, and each private decoder consists of several convolutional blocks.

\begin{figure}[t]
   \centering
   \includegraphics[width=0.8\linewidth]{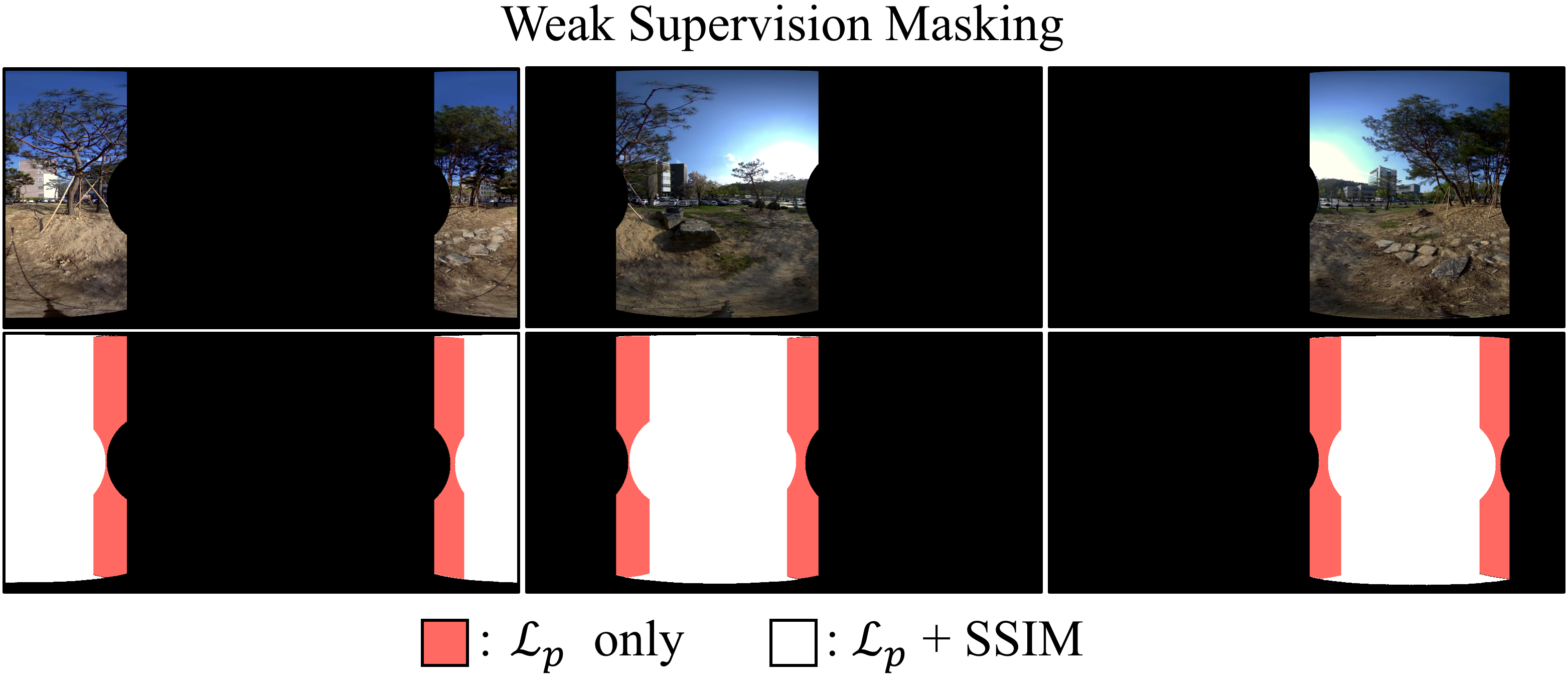}
   \caption{The specific application area of loss functions. Note that SSIM loss is valid only in the white area.}
   \label{fig:validarea}
\end{figure}

\noindent\textbf{Output generation processes.}
First, the color of the $N$ input images with different color tones is corrected by using the estimated pre-color correction map $C_{n}^{pre}$.
Inspired by Zero-DCE~\cite{guo2020zero}, we convert the color intensity values of input images by a monotonic quadratic curve as follows:
\begin{equation}
    \hat{I}_{n} = I_{n} + C_{n}^{pre}I_{n}(1-I_{n}).
    \label{eq:color_correction}
\end{equation}
The color-corrected images $\hat{I}_{n}$ will have color tones harmonized with each other.
Then, each color-corrected input fisheye image is warped to the output pixel grid.
Our output image is in 360$^{\circ}$ ERP format.
Warping is performed by a global warping map $G_{n}$ and a local warping adjustment map $L_{n}$ for each input image.
Since the entire depth layer of the scene cannot be covered with only a single global warping map, the local warping adjustment map is used to supplement it as follows.
\begin{equation}
 U_{n} = G_{n} + \alpha L_{n},
 \label{eq:get-final-flow}
\end{equation}
where $\alpha$ is a balancing factor and set to $0.3$ in our experiment.
Using the final warping map $U_{n}$, color-corrected fisheye images $\hat{I}_{n}$ are warped as
\begin{equation}
 \bar{I}_{n} = \mathbf{warp}(\hat{I}_{n}, U_{n}),
 \label{eq:warping}
\end{equation}
where $\mathbf{warp}(\cdot)$ is a pixel mapping function.
After that, all warped images are weighted and merged to create a panoramic image $P$ as follow:
\begin{equation}
    P=\sum_{n=1}^{N} \bar{I}_{n}  W_{n},
    \label{eq:weighted-sum}
\end{equation}
where $W_{n}$ is a per-pixel weight map for fusing warped images.
Finally, a post-color correction map $C^{post}$ is applied to generate the final panoramic image $O$ as
\begin{equation}
    O = P + C^{post}P(1-P).
    \label{eq:post_color_correction}
\end{equation}
Detailed formulations of the architecture are in the supplementary material.

\subsection{Training}
Learnable parameters of our stitching model $\mathbf{S}(\cdot)$ are trained using the images for weak supervisions generated by the method described in~\Cref{sec:dataset}.
Since genuine GT images do not exist in our settings, we use perceptual loss~\cite{ledig2017photo} instead of pixel-wise loss as follows:
\begin{equation}
    \mathcal{L}_{p}(\theta)  = \sum_{n=1}^{N} \sum_{i=3}^{5} \mathcal{L}_{1}(\phi_{i}(\bar{O}_{n}),\phi_{i}(M_{n}O)),
    \label{eq:loss_perceptual}
\end{equation}
where $\mathcal{L}_{1}(\cdot)$ and $\phi_{i}(\cdot)$ are functions of $L_{1}$ distance and feature extractor at $i$-th maxpooling layer of VGG16~\cite{simonyan2014very}, respectively.
$\bar{O}_{n}$ represents the image for weak supervisions and $M_{n}$ is the mask representing the valid pixels of $\bar{O}_{n}$.
Note that $M_{n}$ is the union of the red and white areas in the bottom row of~\cref{fig:validarea}.
Also, for the consistency in color tone and contrast of the input images, we use SSIM loss as follows:
\begin{equation}
    \mathcal{L}_{SSIM}(\theta) = \sum_{n=1}^{N}[(1-SSIM(\hat{M}\bar{O}_{n},\hat{M}O))],
    \label{eq:loss_ssim}
\end{equation}
where \textit{SSIM}($\cdot$) is a function of the structural similarity~\cite{wang2004ssim}, and $\hat{M}$ is a mask representing non-overlapping regions between the images for weak supervisions.
By using this loss function, our model can harmonize the color tone in the overlapping regions between inputs.
Note that $\hat{M}$ is only white areas in the bottom row of~\cref{fig:validarea}.
Overall loss for training our stitching model is defined as
\begin{equation}
  \mathcal{L}(\theta) = (1-\lambda)\mathcal{L}_{p}(\theta) + \lambda \mathcal{L}_{SSIM}(\theta),
\end{equation}
where $\lambda$ represents the balancing factor between two losses.
We set $\lambda$ to 0.4 in our experiments.
%
\section{Experiments}
\label{sec:experiments}
\subsection{Implementation Details}
%
For the experiments, we use our dataset as well as the CROSS dataset~\cite{li2019cross}.
For our dataset, we use 47,063 sets of images for the training and 1,400 for the test. 
Each training set includes three input fisheye images, three ERP images for weak supervisions, and three masks. 
For the CROSS, we divide the dataset into 1,146 for the training and 128 for the test.
Each set of the CROSS includes two fisheye inputs, and a GT that is pseudo-labeled by SamsungGear.
SamsungGear's MOS obtain the highest in most data, thus we choose it as our pseudo-labeling method.
For both datasets, all images have a resolution of $1024\times512$, and data augmentations such as brightness and tone adjustments are randomly applied during the training.
Our model was trained by Adam Optimizer~\cite{kingma2014adam} with a learning rate of 0.0004. 
The number of epochs for our dataset and the CROSS is set to 20 and 1200, respectively. 
Our method is implemented using Pytorch 1.8.1 with CUDA 11.1 on Ubuntu 18.04.
%
\begin{figure*}[t]
   \centering
   \includegraphics[width=1.0\linewidth]{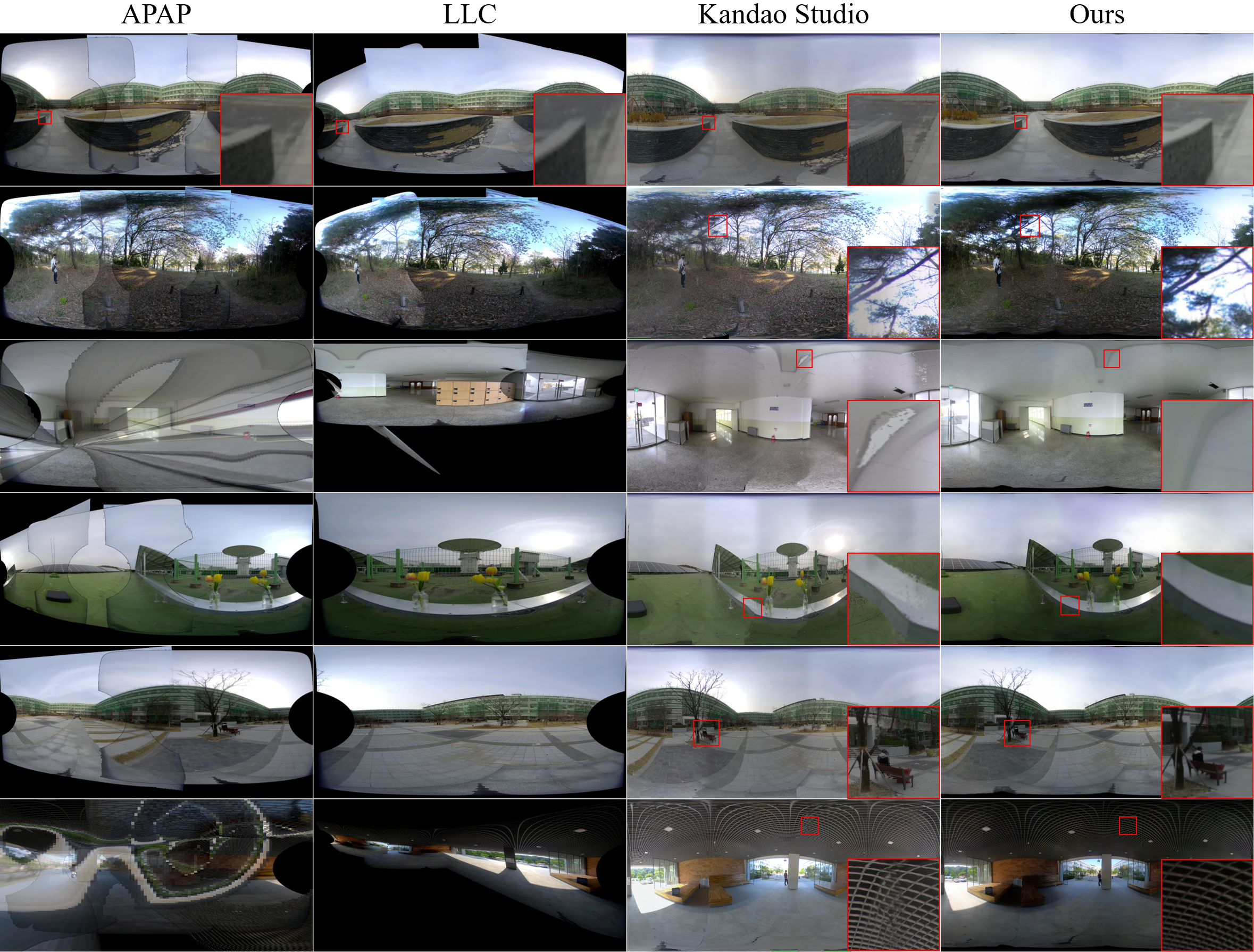}
   \caption{Qualitative comparisons on our dataset. Please refer to the supplementary material for the rest of the test examples.}
   \label{fig:qualitative_others}
\end{figure*}

\begin{table*}
    \addtolength{\tabcolsep}{1.2pt}
    \centering
    \caption{Quantitative result of our 1,400 test dataset. \textbf{bold}: best. Note that Ours$^{\dagger}$ is our model without the post-color correction map.}
    {\scriptsize
    \begin{tabular}{c|cccccccc}
    \toprule
     Metric & APAP~\cite{zaragoza2013projective} & LLC~\cite{jia2021leveraging} & Kandao~\cite{kandao} & Ours$^{\dagger}$(CPU/GPU) & Ours (CPU/GPU) \\ \midrule\midrule
     Time Spent (s) & 8.5887 & 16.5126 & 0.8275 & 1.3010/\textbf{0.0347} & 1.3870/0.0363 \\
     $P_{d}$ $(\downarrow)$ & 6.498 & 6.625 & 5.308 & 2.773 & \textbf{2.731} \\
     LPIPS (Alex) $(\downarrow)$ & 0.647 & 0.722 & 0.266 & 0.122 & \textbf{0.118} \\
     LPIPS (VGG16) $(\downarrow)$ & 0.652 & 0.690 & 0.408 & 0.178 & \textbf{0.175} \\
     SIQE~\cite{madhusudana2019subjective} $(\uparrow)$ & 22.644 & 20.602 & 29.399 & \textbf{39.528} & 37.714 \\
     FID~\cite{heusel2017gans} $(\downarrow)$ & 585.6 & 608.1 & 224.0 & \textbf{132.4} & 140.8 \\
    \bottomrule

\end{tabular}
}
\label{table:quantitative@ours}

\end{table*}
\subsection{Comparisons}
\noindent\textbf{Results on our dataset.}
Since there are no genuine GT images in our datasets, we utilize a perceptual distance $P_{d}$ using VGG16 as an evaluation metric.
The perceptual distance is computed by using making in the same way as in the training step, but there is a difference that unlike in training, the distance is calculated using all five feature maps from five max pooling layers as follows:
%
%
%
%
%
%
%
\begin{equation}
    P_{d}  = \sum_{n=1}^{N} \sum_{i=1}^{5} \mathcal{L}_{1}(\phi_{i}(\bar{O}_{n}),\phi_{i}(M_{n}O)),
    \label{eq:perceptual_distance}
\end{equation}
As a result, $P_{d}$ can evaluate low-level features such as edges.
Since our model is trained in the same way, this evaluation can be unfair. %
Therefore, to compensate for this, we also utilize SIQE~\cite{madhusudana2019subjective}, LPIPS~\cite{zhang2018unreasonable}, and FID~\cite{heusel2017gans} as quantitative evaluation metrics.
%
%
%
%
As for the competition methods, APAP~\cite{zaragoza2013projective}, LLC~\cite{jia2021leveraging}, and Kandao Studio~\cite{kandao} are selected for which the softwares are publicly available.
%
%
%
We use ERP format input images for the APAP and the LLC because they were not developed for fisheye inputs.
Qualitative comparisons are shown in~\figref{fig:qualitative_others}.
Our method produces the most natural, high-quality 360$^{\circ}$ panoramic images without structural distortions and color inconsistency.
In~\Cref{table:quantitative@ours}, there are quantitative comparisons with the existing methods.
%
%
%
%
As ablation studies, we also compare our model without a post-color correction map.
In addition, we measure the average running time per image for all methods.
Note that the running time of the kandao studio includes time for saving a $1920\times960$ image because there are no open-source codes.
%
%
As reported in~\Cref{table:quantitative@ours}, our method performs better and much faster than the existing methods.
However, the results were not significantly different according to a post-color correction map.
%
%
%
Also, as expected, the proposed method using GPU acceleration is much faster than other algorithms, including the commercial kandao studio.
%

\begin{figure*}[t]
   \centering
   \includegraphics[width=1.\linewidth]{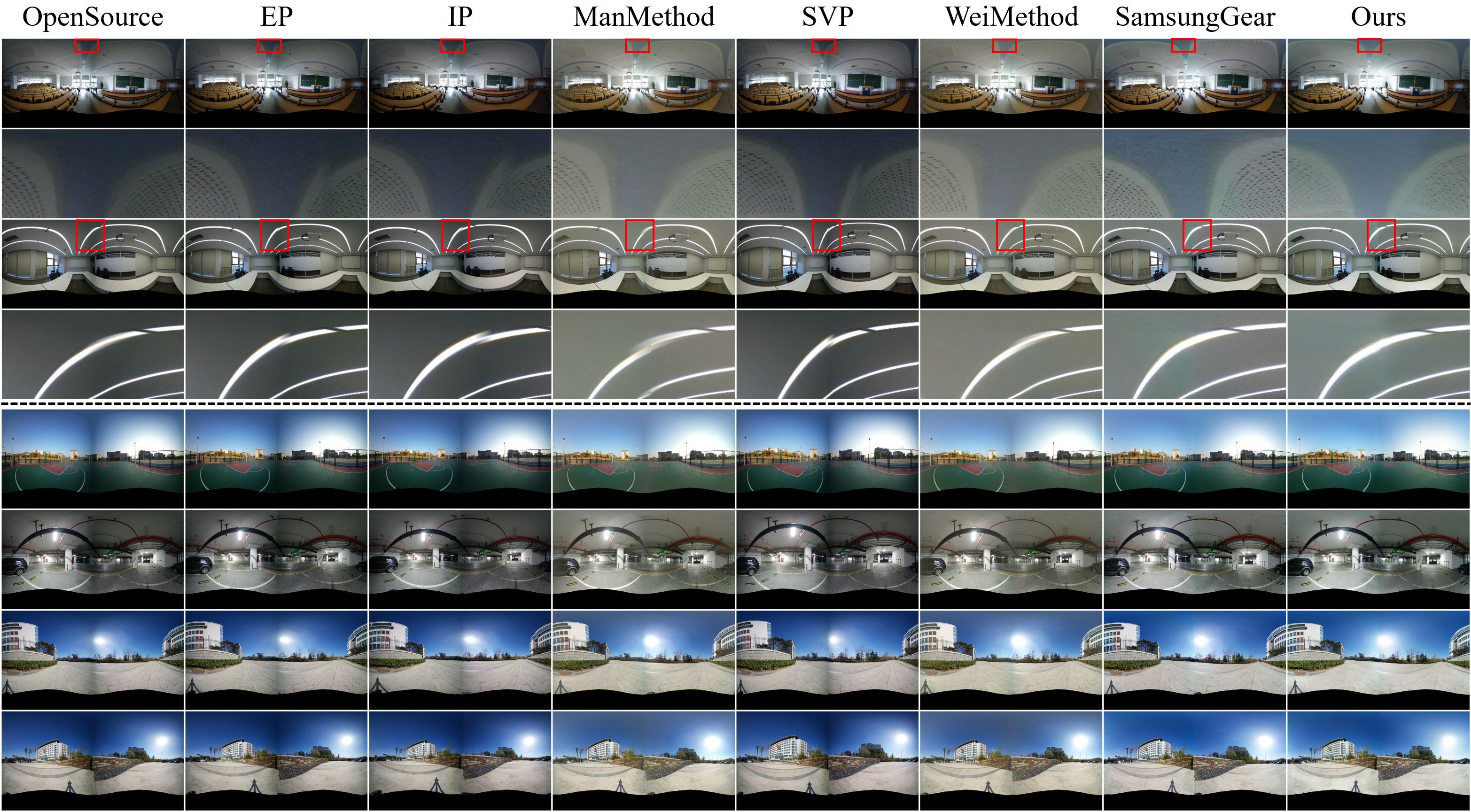}
   \caption{Qualitative results on CROSS dataset. Top: our method well preserves structural patterns compared to existing stitching models. Bottom: our method produces more color-consistent results than other existing methods.}
   \label{fig:qualitative_CROSS}
\end{figure*}
\begin{table*}[t]
    \addtolength{\tabcolsep}{1.2pt}
    \centering
    \caption{Quantitative comparisons on CROSS dataset~\cite{li2019cross}. \textbf{bold}: best.}
    {\scriptsize
    \begin{tabular}{c|ccccccc}
    \toprule
     Metric      & OpenSource~\cite{open-source} & EP~\cite{EP} & IP~\cite{IP} & ManMethod & SVP~\cite{SVP} & WeiMethod~\cite{WeiMethod} & Ours  \\ \midrule\midrule
     PSNR $(\uparrow)$         & 16.417 & 15.908 & 15.177 & 18.943 & 16.110 & 18.730 & \textbf{22.440} \\
     SSIM $(\uparrow)$         & 0.589 & 0.565  & 0.546 & 0.611 & 0.562 & 0.595 & \textbf{0.736} \\
     $P_{d}$ $(\downarrow)$ & 3.31 & 3.55 & 3.79 & 3.12 & 3.69 & 3.23 & \textbf{2.53} \\
    \bottomrule
\end{tabular}

\label{table:quantitative@cross}
}
\end{table*}

\noindent\textbf{Results on the CROSS.}
To validate the versatility of the proposed method, we evaluate the proposed method on the CROSS dataset, which contains pseudo GT $360^{\circ}$ panoramic images as supervisions.
As shown in~\figref{fig:qualitative_CROSS}, our method produces more visually pleasing results compared to the existing stitching methods.
In particular, our results demonstrate robustness to structural distortion and vignetting artifacts.
To measure PSNR, SSIM, and $P_{d}$, we use the pseudo-labeled GT images, because the SamsungGear method obtains the highest MOS in~\cite{li2019cross}.
Note that $M_{n} = 1$ in all pixels because masking is not required.
As reported in~\Cref{table:quantitative@cross}, the proposed model outperforms the existing methods.

\subsection{Ablation Studies}
\begin{figure*}[t]
   \centering
   \includegraphics[width=1.\linewidth]{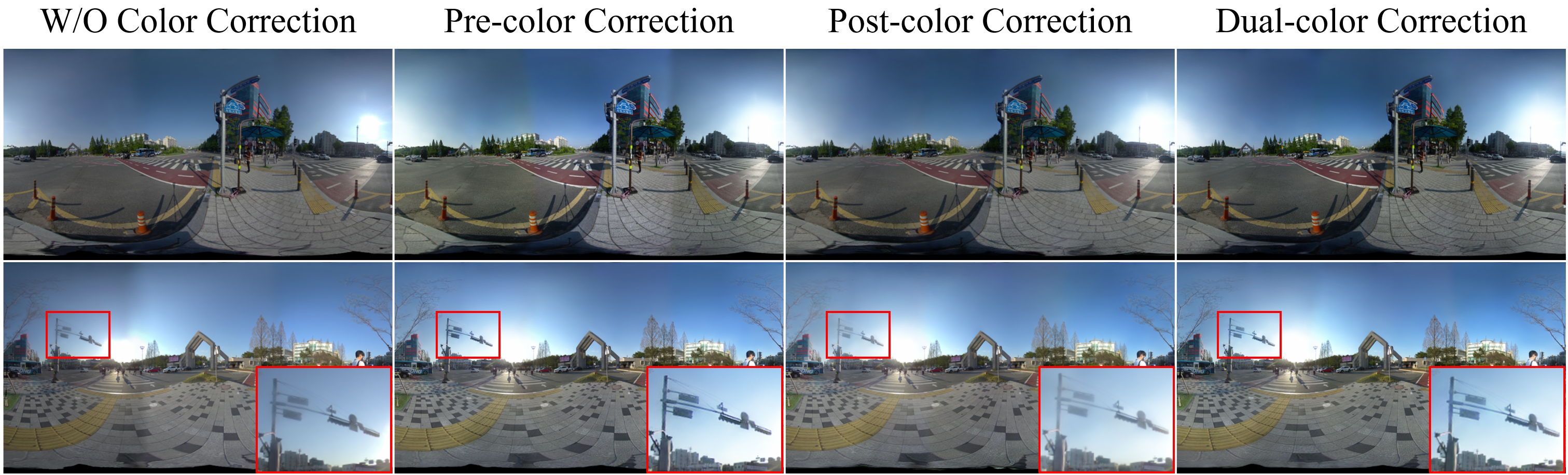}
   \caption{Ablation studies of our model. The w/o color correction and the pre-color correction model have the unpleasant boundaries. The post-color correction model is suffered from fading.
   }
   \label{fig:Ablation}
\end{figure*}

\noindent\textbf{Effects of color correction.}
We conduct experiments depending on whether the pre-color correction map $C_{n}^{pre}$ and the post-color correction map $C^{post}$ are used.
As shown in~\figref{fig:Ablation}, the color tone around boundary lines is inconsistent when only pre-color correction is applied.
In addition, results of only using post-color correction suffer from fading effects as shown in the second row of~\figref{fig:Ablation}.
Overall, the dual-color correction model using both pre-color correction and post-color correction produces the most comfortable results.
\begin{figure}[t]
   \centering
   \includegraphics[width=0.99\linewidth]{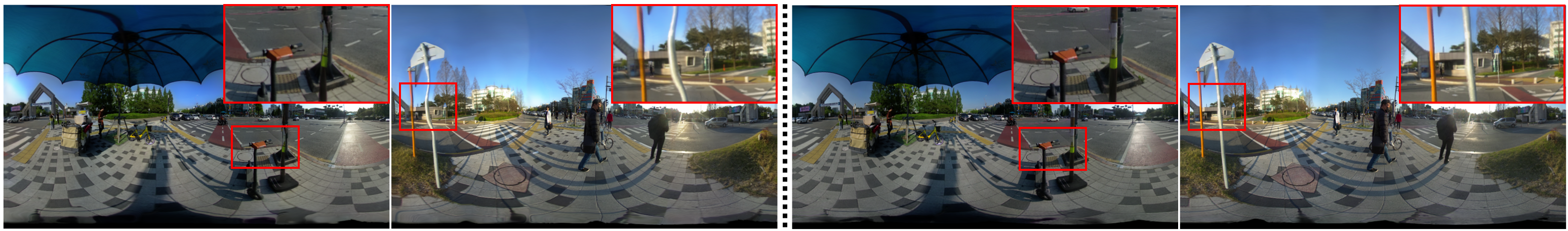}
   \caption{Effects of loss. Utilizing $L_{1}$ loss instead of $\mathcal{L}_{p}$~(left). Ours~(right).}
   \label{fig:Ablation-loss}
\end{figure}

\noindent\textbf{Effects of loss.}
Since the images for weak supervisions have parallax between themselves, it may not be appropriate to use a common pixel-wise regression loss.
Considering this point, we adopt the perceptual loss as in~\eqref{eq:loss_perceptual}.
Therefore, as ablation studies, we train our model with a pixel-wise $L_{1}$ loss instead of~\eqref{eq:loss_perceptual}. 
As shown in~\figref{fig:Ablation-loss}, models trained using $L_{1}$ loss are vulnerable to parallax distortion, which causes noticeable distortion.
%

\begin{table}
    \centering
    \caption{Self-comparisons according to the number of warpings $K$ on our full test dataset (14 sets). The number of epochs is set to 10. \textbf{bold}: best.}
    \addtolength{\tabcolsep}{5pt}
    \renewcommand{\arraystretch}{1.0}
    \begin{tabular}{c|cccccc}
    \toprule
    Metrics & $K$=1 & $K$=2 & $K$=3 & $K$=4 & $K$=5  \\ \midrule\midrule
    $P_{d}$ $(\downarrow)$  & 4.001 & 3.936 & 3.944 & \textbf{3.904} & 3.962 \\
    SIQE $(\uparrow)$ & \textbf{22.89} & 21.16 & 15.28 & 17.89 & 18.29 &  \\
    \bottomrule
\end{tabular}
\label{table:quantitative@ablations}
\end{table}

\noindent\textbf{Effects of the number of warpings.}
Inspired by~\cite{song2021end}, we modified our model to perform multiple $K$ warpings.
However, as shown in~\Cref{table:quantitative@ablations}, multiple warpings do not have a significant effect on the quantitative results.
We guess that it is because our model uses different input and output coordinates from the stitching model in~\cite{song2021end}.
Note that cylindrical coordinates are used in~\cite{song2021end} while our model is operated on fisheye input and ERP output.
Based on the above results, we use the simplest model with $K = 1$ for all experiments.
%

\section{Limitations and Future Works}
Even though our model can be trained without genuine GTs, our research does not take view-free inputs into account.
%
%
%
We believe that subsequent studies based on this paper can be extended to studies on view-free stitching.
Another promising future work is video stitching to cover dynamic scenes.
Although the proposed method is developed for a static scene, it can be extended to video, and we believe that temporal artifacts such as waving effects can be solved by temporal consistency loss as in~\cite{lai2019video}.

\section{Conclusion}
In this paper, we present a weakly supervised method for training the real-world stitching model.
Our model takes multiple fisheye images as inputs and generates a $360^{\circ}$ panorama image.
For training, we generate images of weak supervisions and utilize them for perceptual and SSIM losses.
We verify the proposed method on our stitching dataset as well as the CROSS dataset.
Through the various experiments, we demonstrate superior stitching performance over existing methods.
In particular, it is more robust to structural artifacts and color inconsistency problems compared to existing methods.

\section*{Acknowledgement}
This work was supported by Institute of Information $\&$ Communications Technology Planning $\&$ Evaluation (IITP) grant funded by the Korea government(MSIT) (No. 2018-0-00207, Immersive Media Research Laboratory) and the National Research Foundation of Korea(NRF) grant funded by the Korea government(MSIT) (No.2021R1A4A1032580, No.2022R1C1C1009334).

%
%
\bibliographystyle{splncs04}
\bibliography{egbib}
\end{document}